\def\year{2020}\relax
\documentclass[letterpaper]{article} 
\usepackage{aaai20}  
\usepackage{times}  
\usepackage{helvet} 
\usepackage{courier}  
\usepackage[hyphens]{url}  
\usepackage{graphicx} 
\usepackage{graphicx}  
\usepackage{latexsym}
\usepackage{epstopdf}
\usepackage{multirow}
\usepackage{graphicx}
\usepackage{url}
\usepackage{amsmath}
\usepackage{latexsym}
\usepackage{amsfonts}

\newcommand{\citet}[1]{\citeauthor{#1} \shortcite{#1}}
\newcommand{\citep}{\cite}

\newcommand{\wordvec}{\texttt{word2vec}}
\newcommand{\fastext}{\texttt{FastText}}
\newcommand{\glove}{\texttt{\textsc{G}lo\textsc{V}e}}
\newcommand{\elmo}{\texttt{\textsc{ELMo}}}
\newcommand{\bert}{\texttt{\textsc{BERT}}}

\title{Parsing as Pretraining}

\author{David Vilares,\textsuperscript{\rm 1} Michalina Strzyz,\textsuperscript{\rm 1} Anders S{\o}gaard,\textsuperscript{\rm 2,3} Carlos G{\'o}mez-Rodr{\'i}guez\textsuperscript{\rm 1} \\  
\textsuperscript{\rm 1} Universidade da Coru{\~n}a, CITIC,  Ciencias de la Computaci{\'o}n y Tecnolog{\'i}as de la Informaci{\'o}n (CC\&TI), A Coru{\~n}a, Spain\\
\textsuperscript{\rm 2} University of Copenhagen, Department of Computer Science, Copenhagen, Denmark\\ \textsuperscript{\rm 3} Google Research, Berlin, Germany\\
david.vilares@udc.es, michalina.strzyz@udc.es, soegaard@di.ku.dk, carlos.gomez@udc.es
}

\begin{document}
\maketitle

\begin{abstract}

Recent analyses suggest that encoders pretrained for language modeling capture certain morpho-syntactic structure. However, probing frameworks for word vectors still do not report results on standard setups such as constituent and dependency parsing. This paper addresses this problem and does full parsing (on English) relying \emph{only} on pretraining architectures -- and {\em no} decoding. We first cast constituent and dependency parsing as sequence tagging. We then use a single feed-forward layer to directly map word vectors to labels that encode a linearized tree. This is used to: (i) see how far we can reach on syntax modelling with just pretrained encoders, and (ii) shed some light about the syntax-sensitivity of different word vectors (by freezing the weights of the pretraining network during training). For evaluation, we use bracketing F1-score and \textsc{las}, and analyze in-depth differences across representations for span lengths and dependency displacements. The overall results surpass existing sequence tagging parsers on the \textsc{ptb} (93.5\%) and end-to-end \textsc{en-ewt ud} (78.8\%). 

\end{abstract}

\section{Introduction}

Traditionally, natural language processing (\textsc{nlp}) models  represented input sentences using one-hot vectors, together with weighting schemes such as \textsc{tf-idf}. Such vectors can encode shallow linguistic information, like term frequency or local context if n-grams are allowed. However, they cannot capture complex linguistic structure due to the inability to consider non-local context, their orderless nature, and the curse of dimensionality.

This paradigm has however become obsolete for many NLP tasks in favour of  continuous vector representations, also known as word embeddings. The idea is to encode words as low-dimensional vectors ($\vec{v} \in \mathbb{R}^n$), under the premise that words with similar context should have similar vectors -- which seemingly better capture semantic and morpho-syntactic information. In practice, these architectures have been widely adopted because they have not only made it possible to obtain more accurate models but also conceptually simpler ones, reducing the number of features required to obtain state-of-the-art results. 

In parsing, two research paths have arisen with respect to word vectors:
(i) whether and to what extent pretraining architectures can offer help creating parsers which avoid the need for dozens of hand-crafted structural features, \emph{ad-hoc} parsing algorithms, or task-specific decoders; and (ii) how to explain what sort of syntactic phenomena are encoded in such pretraining encoders.
In related work, to test (i) it is common to rely on ablation studies to estimate the impact of  removing features, or to further contextualize word embeddings with powerful, task-specific neural networks to show that richer linguistic contextualization translates into better performance. But to the best of our knowledge, there is no work that has tried to do (full) parsing relying \emph{uniquely} on word embeddings, i.e. no features beyond words, no parsing algorithms, and no  task-specific decoders. To test (ii), the most common probing framework consists in using models with limited expression (e.g. feed-forward networks on top of the word vectors) to solve tasks that can give us insights about the linguistic information that word vectors can encode \cite{tenney2018you}. However, these recent studies do not provide results on full parsing setups, but instead on simplified versions, which sometimes are even limited to analyzing capabilities on specific syntactic phenomena.

\paragraph{Contribution}  
In this paper, we try to give an answer to these questions using a unified framework. Our approach consists in casting both (full) constituent and dependency parsing as pretraining from language modelling.\footnote{In this paper, we limit our analysis to English.} To do so, we first reduce constituent and dependency parsing to sequence labeling. Then, under this paradigm we can directly map, through a single feed-forward layer, a sequence of word embeddings of length $n$ into an output sequence (also of length $n$) that encodes a linearized tree. The novelty of the paper is twofold: (i) we explore to what extent it is possible to do parsing relying {\em only}~on pretrained encoders, (ii) we shed light on the syntactic abilities of existing encoders.

\section{Related work}\label{section-related-work}

We now review previous work on the two research directions that we will be exploring in our paper.

\subsection{Word vectors for simpler parsing}

Syntactic parsers traditionally represented input sentences as one-hot vectors of discrete features. Beyond words, these included part-of-speech (PoS) tags, morphological features and dozens of hand-crafted features. For instance, in transition-based dependency parsers it was common to refer to daughter or grand-daughter features of a given term in the stack or the buffer, in order to provide these models with more contextual information  \cite{zhang2011transition}. We will be referring to these features as \emph{structural features}.

\citet{chen2014fast} were one of the first to use word vectors to train a transition-based dependency parser using a feed-forward network. They showed that their model performed comparably to previous parsers, while requiring fewer structural features. 
Later, \citet{kiperwasser2016simple} demonstrated that replacing \citeauthor{chen2014fast}'s feed-forward network with bidirectional long short-term memory networks (\textsc{bilstm}s) \cite{hochreiter1997long} led to more accurate parsers that at the same time required even fewer features. Following this trend, \citet{shi2017fast} proposed a minimal feature set, by taking advantage of \textsc{bilstm}s and dynamic programming algorithms. Furthermore, this redundancy of structural features in neural parsers was empirically demonstrated by \citet{falenska-kuhn-2019-non}. However, in addition to small sets of hand-crafted features, these approaches relied on \emph{ad-hoc} parsing algorithms. In this vein, recent research has showed that task-specific sequence-to-sequence and sequence labeling decoders suffice to perform competitively even without small sets of structural features nor parsing algorithms \cite{LiSeq2seq2018,StrzyzViable2019}.

For constituent parsing, the tendency has run parallel. Transition-based systems \cite{zhu2013fast} used templates of hand-crafted features suiting the task at hand. More recent research has shown that when using word embeddings and neural networks such templates can be simplified \cite{dyer2016recurrent} or even ignored \cite{kitaev2018constituency,kitaev2018BERT}. Finally, it has been proved that transition-based or chart-based parsers are not required to do constituent parsing, and that task-specific neural decoders for sequence-to-sequence and sequence labeling suffice \cite{vinyals2015grammar,gomez2018constituent}.

\subsection{Probing syntax-sensitivity of word vectors}

These improvements in parsing have raised the question of whether and to what extent pretraining architectures capture syntax-sensitive phenomena. For example, \citet{mikolov2013distributed} and \citet{pennington2014glove} already discussed the syntax-sensitivity of \wordvec\ and \glove\ vectors, evaluating them on syntactic word analogy (e.g. `seat is to seating as breath is to \emph{x}'). However this only provides indirect and shallow information about 
what syntactic information such vectors accommodate.

State-of-the-art pretrained encoders such as \elmo~\cite{peters2018deep} or \bert~\cite{devlin2018bert}, trained with a language modeling objective and self-supervision, seemingly encode some syntactic properties too. \citet{goldberg2019assessing} discusses this precisely using \bert, and reports results on subject-verb agreement tasks (e.g. `The \emph{dog} of my uncle \emph{eats}' vs `The \emph{dog} of my uncles \emph{eat}'). He used similar methods to the ones employed to assess how recurrent neural networks (\textsc{rnn}) capture structure. For example, \citet{linzen2016assessing} studied how \textsc{lstm}s perform on subject-verb agreement, and trained the network using different objectives to know whether it predicted the grammatical number of the next word. \citet{gulordava2018colorless} additionally incorporated the concept of `colorless green ideas' \cite{chomsky1957syntactic}, i.e. they replaced content words with random terms 
with the same morphological information, 
to force the model to attend to syntactic patterns and not words.

Another common strategy consists in analyzing the capabilities of word vectors using models with limited expression, i.e. simple models such as n-layer feed-forward networks that take word vectors as input and are used to solve structured prediction tasks in NLP. This is the angle taken by \citet{tenney2018you}, who evaluated contextualized embeddings on problems such as part-of-speech tagging or named-entity labeling. They also included \emph{simplified and partial} versions of constituent and dependency parsing. They ask the model to predict the type of phrase of a span or the dependency type between two specific words, i.e., not to predict the full syntactic structure for a sentence. 
In a similar fashion, \citet{liu-etal-2019-linguistic} analyze syntactic properties of deep contextualized word vectors using shallow syntactic tasks, such as CCG supertagging \cite{clark2002supertagging}. 

In a different vein, \citet{hewitt2019structural} proposed a structural probe to evaluate whether syntax trees are embedded in a linear transformation of an \elmo\ and \bert\ word representation space. Although their study does not support full parsing analysis either, it reports partial metrics.\footnote{These partial metrics refer to analysis such as \emph{undirected} Unlabeled Attachment Score (UUAS) as well as the average Spearman correlation of true to predicted distances.}

\section{Parsing only with word vectors}

Nevertheless, it remains an open question how much of the workload these encoders can take off a parser's shoulders. To try to answer this, we build on top of previous work reducing parsing to sequence tagging as well as on using models with limited capacity. The goal is to bring to bear what pretrained encoders have learned from distributional evidence alone.

\paragraph{Notation} We will be denoting a raw input sequence by $w=[w_0,w_1,...,w_{|w|}]$, with $w_i \in V$ and mark vectors and matrices with arrows (e.g. $\vec{v}$ and $\vec{W}$).

\subsection{Parsing as sequence labeling}

Sequence labeling is a structured prediction task where given an input sequence, $w_{|w|}$, the goal is to generate one output label for every $w_i$. Part-of-speech tagging, chunking or named-entity recognition are archetype tasks of this type of problem. In this vein, recent work has demonstrated that is possible to design reductions to address full constituent and dependency parsing as sequence labeling too.\footnote{Whether sequence labeling is the most adequate paradigm, in terms of performance, for obtaining syntactic representations (in contrast to algorithms) is a fair question. That said, the aim of this work is not to outperform other paradigms, but to find a way 
to estimate the `amount of syntax' that is encoded in embeddings. To achieve this, we consider that sequence labeling with limited expressivity is a natural probing framework for syntax-sensitivity.}

\subsubsection{Constituent parsing as sequence labeling}

\begin{figure}[hbtp]
\centering
\includegraphics[width=0.65\columnwidth]{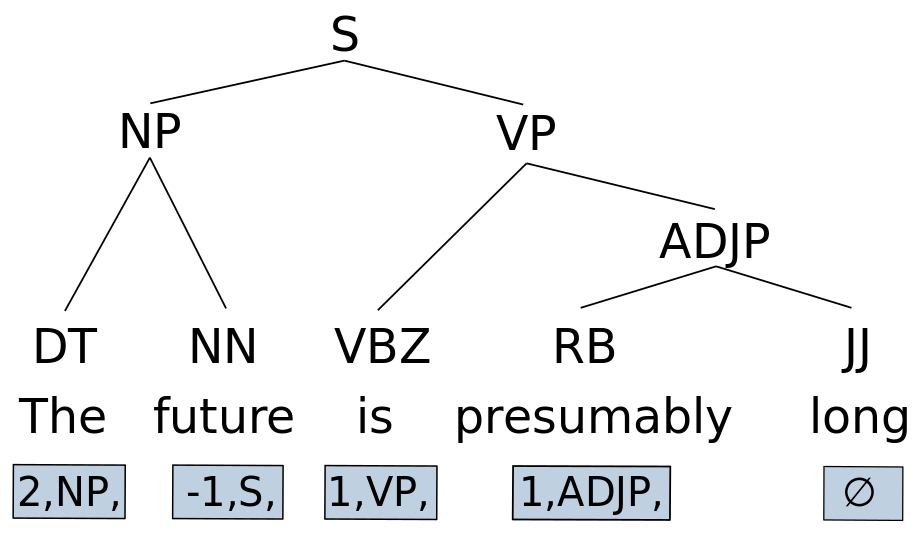}
\caption{\label{f-seq-lab-example} A linearized constituent tree according to \citet{gomez2018constituent}. For example, $n_{\textit{`The'}}=2$ since `The' and `future' have in common the top two levels in the tree (and there is no previous $n$), $c_{\textit{`The'}}$=NP because that is the non-terminal symbol shared at that lowest common ancestor, and $u_{\textit{`The'}}=\emptyset$ as there is no unary branch for `The'. If we move one step forward, for `future' we have $n_{\textit{`future'}}=-1$ (in terms of the variation with respect to the previous timestep), since `future' and `is' just share the top level of the tree, i.e. one level less than for $n_{\textit{`the'}}$. }
\end{figure}

\citet{gomez2018constituent} reduced constituent parsing to a pure sequence labeling task, defining a linearization function $\Phi_{|w|} : T_{|w|} \rightarrow L^{|w|} $ to map the constituent tree of an input sentence $w$=$[w_0,w_1,...,w_{|w|}]$ into a sequence of $|w|$ labels, i.e. they proposed a method to establish a one-to-one correspondence between words and labels that encode a syntactic representation for each word with enough information to encode the full constituent tree of the sentence. In particular, each label $l_i \in L$ is a 3-tuple ($n_i$,$c_i$,$u_i$) where:\footnote{The index is omitted when not needed.}
\begin{itemize}
    \item $n_i$ encodes the number of tree levels in common between $w_i$ and $w_{i+1}$ (computed as the relative variation with respect to $n_{i-1}$). 
    \item $c_i$ encodes the lowest non-terminal symbol shared between those two words.
    \item $u_i$ encodes the leaf unary branch located at $w_i$ (if any). 
\end{itemize}
Figure \ref{f-seq-lab-example} illustrates the encoding with an example.

\paragraph{Postprocessing} The linearization function is complete and injective, but not surjective, i.e. postprocessing needs to be applied in order to ensure the validness of a predicted output tree. We follow the original authors' strategy to solve the two potential sources of incongruities:
\begin{enumerate}
    \item conflicting non-terminals, i.e. a nonterminal $c$ can be the lowest common ancestor of more than two pairs of contiguous words $(w_i,w_j)$ with $c_i \neq c_j$. In such case, we ignore all but the first prediction.
    \item empty intermediate unary branches, i.e. decoding the sequence of $n_i$'s might generate a predicted tree where some intermediate unary branches are not assigned any non-terminal. If so, we simply delete that empty level.
\end{enumerate}

\subsubsection{Dependency parsing as sequence labeling}

\begin{figure}[hbtp]
\centering
\includegraphics[width=0.80\columnwidth]{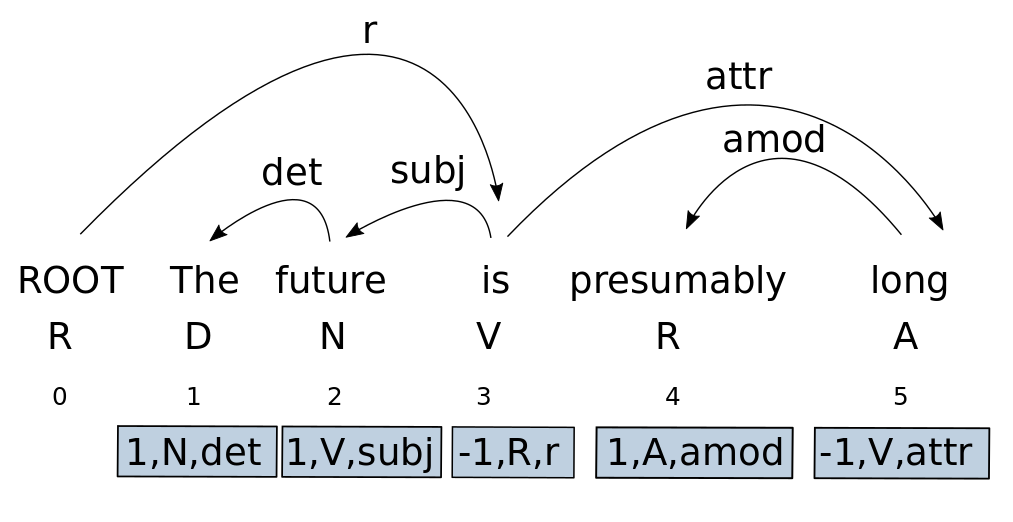}
\caption{\label{f-seq-dep-lab-example} A linearized dependency tree according to the PoS tag-based encoding used in \citet{StrzyzViable2019}. For instance, $(o_{\textit{`The'}},p_{\textit{`The'}})=(1,N)$ which means that the head of `The' is the first word to the left whose part-of-speech tag is N, i.e. `future'; and $d_{\textit{`The'}}$= det, which denotes the syntactic relationship existing between the head `future' and the dependent `The'.}
\end{figure}

In a similar fashion to \citet{gomez2018constituent}, \citet{StrzyzViable2019} define a function to encode a dependency tree as a sequence of labels, i.e. $\Upsilon_{|w|}: T_{d,|w|} \rightarrow L_{d}^{|w|}$. Let $r_i \in L_{d}$ be a particular label that encodes the head term of $w_i$, they also represent it as a 3-tuple $(o_i,p_i,d_i)$ where:\footnote{Note that a dependency tree can be trivially encoded as a sequence of labels using a na\"ive positional encoding that uses the word absolute index and the dependency type, but in previous work this did not lead to robust results.}
\begin{itemize}
\item The pair $(o_i,p_i)$ encodes the index of the head term. Instead of using the absolute index of the head term as a part of the label, the head of $w_i$ is represented as the $o_i$th closest word to the right with PoS tag $p_i$ if $o_i > 0$, and the $-o_i$th closest word to the left with PoS tag $p_i$, if $o_i <0$.
\item $d_i$ denotes the dependency relation between the head and the dependent.
\end{itemize}
Figure \ref{f-seq-dep-lab-example} illustrates the encoding with an example.

\paragraph{Postprocessing} We ensure that the predicted dependency tree is acyclic and single-headed:
\begin{enumerate}
    \item If no token is selected as syntactic root (by setting its head to be index $0$, which we use as a dummy root, as seen in Figure \ref{f-seq-dep-lab-example}), take as root the first one with d=\emph{root}, or  (if none exists) the first token. If multiple roots have been assigned, the first one is considered as the only root, and the rest become children of that one.
    \item If a token has an invalid head index, attach it to the real syntactic root.
    \item If there are cycles, the first token involved is assigned to the real root. The process is repeated until the tree is acyclic.
\end{enumerate}

\section{Models with limited expression}

Parsing as sequence labeling can be used to set up a direct, one-to-one mapping from words to some sort of syntactic labels that help establish conclusions about the syntactic properties that such word vectors accommodate. 
However, previous work on parsing as sequence labeling did not exploit this paradigm to study this. Instead, they simply trained different task-specific decoders such as \textsc{bilstm}s, which can potentially mask poor syntactic abilities of word representations.

We propose to remove any task-specific decoder and instead directly map $n$ word vectors (extracted from a given pretrained architecture) to $n$ syntactic labels that encode the tree (using a single feed-forward layer, the simplest mapping strategy). Such architecture can be used to study the two research questions we address. First, it can be used to explore how far we can reach relying only on pretraining architectures. At the same time, these models can be used to probe syntax-sensitivity of continuous vector representations. More particularly, they minimize the risk of neural architectures, training objectives or specific parsing algorithms implicitly hiding, modifying and biasing the syntactic abilities captured by word representations during pretraining.
This is in line with \citet{tenney2018you}, who explore properties of contextualized 
vectors by fixing their representations and training a 2-layer perceptron for certain tasks. The aim was to have a model with limited expression, to focus on the information that is directly represented in the vectors. However, when evaluating syntax-sensitivity, \citeauthor{tenney2018you} relied on a simplified and partial version of constituent and dependency parsing, as mentioned in the related work section.
It is also worth remarking that we take a more extreme approach, and use just a single feed-forward layer.

\subsection{Model architecture}\label{section-models}
\begin{figure}[t]
\centering
\includegraphics[width=1.\columnwidth]{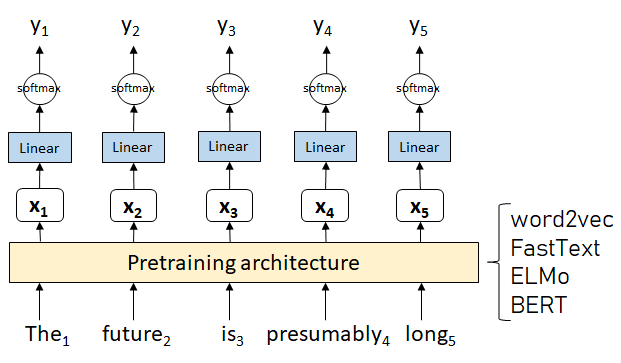}
\caption{\label{f-high-level-architecture} High level architecture of parsing as pretraining}
\end{figure}

We denote the output of a pretrained encoder by $\vec{x}$ = $[\vec{x}_0,\vec{x}_1,...,\vec{x}_{|x|}]$, where $\vec{x}_i$ is the vector for the word indexed at $i$. PoS tags and character embeddings are not used to train the model.\footnote{Note that for dependency parsing we have predicted PoS tags separately to rebuild the tree in CoNLL format, as the output labels contain the PoS tag of the head term. We understand this could lead to some latent learning of PoS tagging information.}
Our architecture is depicted in Figure \ref{f-high-level-architecture}: we use a feed-forward layer
on top of the pretrained encoder to predict each label, $y_i$, followed by a softmax activation function:

\begin{equation}
P(y=j|\vec{x}_i) = \mathit{softmax}(\vec{W} \cdot \vec{x}_i + \vec{b}) = \frac{e^{\vec{W}_j \cdot \vec{x}_i}}{\sum_{k}^{K}{e^{\vec{W}_k \cdot \vec{x}_i}}}
\end{equation}

The model is optimized using categorical cross-entropy:

\begin{equation}
    \mathcal{L} = -\sum{log(P(y|\vec{x}_i))}
\end{equation}

We allow to freeze or fine-tune the word vectors during training, and refer to the models
as \texttt{ff} (feed-forward) and \texttt{ff-ft}, respectively. Freezing word representations aims to not adapt the syntax-sensitivity inherently encoded. 
The goal of fine-tuning is to test the best performance that we can achieve using just pretraining networks.

\subsubsection{Models with task-specific decoders} We also build models using task-specific decoders. More particularly, we use for this setup 2-stacked \textsc{bilstm}s. This is similar to \citet{gomez2018constituent,StrzyzViable2019}. This has a secondary goal: to further illustrate if the tendencies observed with the models with limited expression remain, and thus to provide a more complete evaluation on how choices of embeddings interact with the presence or absence of task-specific decoding. We again will freeze and fine-tune the pretraining architectures, and will refer these models as \texttt{lstm} and \texttt{lstm-ft}.

\subsection{Pretrained encoders}\label{section-pretrained-architectures}

We test both \emph{precomputed} (lookup tables that map words to vectors), and \emph{contextualized} representations (where each word vector is dependent on the context and generated by a pretrained neural network). In this work, we test:

\begin{itemize}
    \item Random embeddings: Uniformly distributed, in the interval $[-\sqrt(3.0) / d, \sqrt(3.0) / d]$, where $d=300$ is the dimension of the embedding. We tried other dimensions corresponding to the size of other tested methods, but we did not observe significant differences. We consider this can roughly be seen as a suitable baseline 
    to know whether the tested embedding methods learn syntax above expectation by chance.
    \item {Skip-gram \wordvec}~\cite{mikolov2013distributed}. The approach learns to predict the window context based on the word, learning to generate the representation for that word in the process.\footnote{For \wordvec, we use GoogleNews-vectors-negative300 (\url{https://code.google.com/archive/p/word2vec/)}}
    \item Structured skip-gram \wordvec~\cite{ling2015two}. A variant of the standard \wordvec\ that is sensitive to word order, keeping separate matrices to learn to predict the word at each position in the window context.\footnote{For structured skip-gram \wordvec, we use the vectors at \url{https://github.com/clab/lstm-parser/blob/master/README.md}}
    \item Cbow-based {\fastext}~\cite{bojanowski2017enriching}. A cbow \wordvec\ extension. It tries to predict a word based on its context, learning to generate its representation in the process.\footnote{\fastext\ can consider subword level information. In early experiments we tried both wiki-news-300d-1M-subword and wiki-news-300d-1M pretrained vectors, choosing the latter because they behaved better (\url{https://fasttext.cc/docs/en/english-vectors.html}).}

    \item {\glove}~\cite{pennington2014glove}. It creates precomputed word vectors combining matrix factorization with local window methods.\footnote{For \glove, we use the glove.840B.300 vectors (\url{https://nlp.stanford.edu/projects/glove/})}
 
    \item {\elmo}~\cite{peters2018deep}. Each word vector is a weighted average: (i) a context-independent vector computed through a character convolutional network, (ii) an output vector from a 2-layer left-to-right \textsc{lstm}, and (iii) and output vector from a 2-layer right-to-left \textsc{lstm}. Following the canonical paper, we let the fine-tuned models learn a linear combination of these representations, but will freeze the language modeling layers.\footnote{\elmo\ can be downloaded from \url{https://allennlp.org/elmo}}

\item {\bert}~\cite{devlin2018bert}. Uses a Transformer to generate the output word vectors.
As the Transformer purely relies on attention mechanisms, the positional information is encoded through positional embeddings, which poses an interesting difference with respect to \elmo, where such information is inherently encoded through the recurrent network.\footnote{For \bert, we used \texttt{bert-base-cased} (\url{https://github.com/google-research/bert})}
\end{itemize}

When available, we chose 300-dimensional vectors, but \elmo\ vectors have 1024 dimensions, \bert\ 768, and the \citet{ling2015two} ones just 100. Despite  this, we stuck to available pretrained models since we feel the experiments are more useful if performed with the standard models used in \textsc{nlp}. Also, it lies out of the standard computational capacity to train \bert\ and \elmo\ models from scratch to force them to have the same number of dimensions.

We build on top of the framework by \citet{yang2018ncrf++} to run all vectors under our setup, except \bert, for which we use a pytorch wrapper and its hyperparameters.\footnote{https://github.com/huggingface/pytorch-pretrained-BERT. For \texttt{ff}/\texttt{lstm}, the learning rate was set to 5e-4.}

\section{Experiments}

The source code is accessible at https://github.com/aghie/parsing-as-pretraining. 

\subsection{Corpora} We use the English Penn Treebank (\textsc{ptb}) \cite{marcus1993building} for evaluation on constituent parsing, and the \textsc{en-ewt} \textsc{ud} treebank (v2.2) for dependency parsing   \cite{UD2.1}. To train our sequence labeling models, we add dummy beggining- and end-of-sentence tokens, similarly to previous work on parsing as sequence labeling. The labels are predicted atomically. 

\subsection{Metrics} For constituents, we use labeled bracketing F1-score and the \textsc{collins.prm} parameter file. We also break down the results according to different span lengths. For dependencies, we use \textsc{uas}\footnote{Unlabeled Attachment Score: Percentage of relations for which the head has been assigned correctly.} and \textsc{las}\footnote{Labeled Attachment Score: Percentage of relations for which the head and the dependency type have been assigned correctly.}.
We use the \textsc{en-ewt ud} treebank with predicted segmentation by UDPipe \cite{straka:2018:K18-2}, for comparison against related work. We also compute \emph{dependency displacements}, i.e., signed distances between the head and the dependent terms (where the dependency type is predicted correctly too). 

\subsection{Results and discussion}

Table \ref{table-const-f1-score} shows the  F1-scores for constituent parsing on the \textsc{ptb}, and both \textsc{uas} and \textsc{las} scores for dependency parsing on the \textsc{en-ewt ud}; evaluating both models where the pretraining network is frozen (\texttt{ff}) and fine-tuned (\texttt{ff-ft}) during training. For a better understanding of how the tendencies of these results can be influenced by a task-specific decoder, Table \ref{table-lstm-results} replicates the same set of experiments using instead a 2-layered \textsc{bilstm} decoder, allowing to freeze (\texttt{lstm}) and fine-tune (\texttt{lstm-ft}) the pretraining weights.

\begin{table}[hbtp]
\tabcolsep=0.12cm
\centering
\small{
\begin{tabular}{l|cc|cccc}

\hline
& \multicolumn{2}{c|}{\bf \textsc{ptb}}&\multicolumn{4}{c}{\bf \textsc{en\_ewt}}\\
\bf Vectors & \bf \texttt{ff} & \bf \texttt{ff-ft} &\multicolumn{2}{c}{\bf \texttt{ff}} & \multicolumn{2}{c}{\bf \texttt{ff-ft}} \\
&\bf \textsc{f1}&\bf \textsc{f1}&\bf \textsc{uas} & \bf \textsc{las} & \bf \textsc{uas} & \bf \textsc{las}  \\
\hline
Random&32.9&42.8&37.3&30.6&46.4&39.7\\
\glove &37.9&42.9&44.7&38.0&47.1&40.3\\
Struct.\wordvec&40.0&43.2&45.4&37.8&47.0&40.2\\
\wordvec&39.1&43.5&45.6&37.9&46.8&40.2\\
\fastext&41.4&43.0&46.6&39.0&47.5&40.3\\
\elmo&69.7&75.8&65.2&60.3&67.6&62.4\\
\bert&\bf 78.2&\bf 93.5&\bf 68.1&\bf 63.0&\bf 81.0&\bf 78.8\\
\hline
\end{tabular}
}
\caption{Labeled F1-score on the \textsc{ptb} test set, and {\textsc{uas/las} on the \textsc{en-ewt ud} test set (with predicted segmentation)}\label{table-const-f1-score}}
\end{table}

\begin{table}[hbtp]
\tabcolsep=0.12cm
\centering
\small{
\begin{tabular}{l|cc|cccc}
\hline
& \multicolumn{2}{c|}{\bf \textsc{ptb}}&\multicolumn{4}{c}{\bf \textsc{en\_ewt}}\\
\bf Vectors & \bf \texttt{lstm} & \bf \texttt{lstm-ft} &\multicolumn{2}{c}{\bf \texttt{lstm}} & \multicolumn{2}{c}{\bf \texttt{lstm-ft}} \\
&\bf \textsc{f1}&\bf \textsc{f1}&\bf \textsc{uas} & \bf \textsc{las} & \bf \textsc{uas} & \bf \textsc{las}  \\
\hline
Random&80.7&88.0&62.3&56.5&73.2&69.7\\
\glove&89.2&89.5&74.3&71.2&75.6&72.5\\
Struct.\wordvec&89.0&89.5&73.2&69.5&74.7&71.5\\
\wordvec&89.1&89.6&72.1&68.5&74.7&71.5\\
\fastext&89.0&89.5&73.2&69.9&74.6&71.5\\
\elmo&\bf 92.5&92.7&\bf 78.8&\bf 76.5&79.5&77.4\\
\bert&92.2&\bf 93.7&78.4&75.7&\bf 81.1&\bf 79.1\\
\hline
\end{tabular}
}
\caption{Same scores reported in Table \ref{table-const-f1-score} on the \textsc{ptb} and \textsc{en-ewt ud}, but using instead a (task-specific) 2-layer \textsc{bilstm} decoder (freezing and fine-tuning the pretraining network).}\label{table-lstm-results}
\end{table}

\subsection{Parsing with pretraining architectures} 

We first discuss the performance of the \texttt{ff} and \texttt{ff-ft} models.
Among the precomputed word vectors, window-based approaches perform slightly better than alternatives such as \glove, but they all do clearly surpass the control method (random embeddings). With respect to contextualized ones, for \texttt{ff}, both \elmo\ and \bert\ get a decent sense of phrase structures, with \bert\ slightly superior.
For the \texttt{ff-ft} models differences are larger: while \elmo\ is well behind a robust model (we believe this could be due to keeping the \textsc{bilm} weights fixed as done by default in \cite{peters2018deep}), \bert\ \texttt{ff-ft} performs competitively. From the results in Table \ref{table-lstm-results} using task-specific decoders, we observe that even if the tendency remains, the differences across different vectors are smaller. This is due to the capacity of these decoders to obtain richer contextualized representations for the target task and to mask poor syntax capabilities of the input, even when the pretrained network is kept frozen.

Tables \ref{table-SOTA-const} and \ref{table-SOTA-dep} compare the \bert-based models without task-specific decoders against the state of the art for constituent and dependency parsing, respectively. In this context, it is worth remarking that all our models only use words as features. For constituents, \bert\ obtains a 93.5\% F1-score vs the 95.8\% reported by \citet{kitaev2018BERT} and \citet{zhou-zhao-2019-head}, which are the current best performing models and use \bert\ representations as input (together with PoS tag embedddings, and in the case of \citeauthor{zhou-zhao-2019-head} also char embeddings). Surprisingly, the performance of \bert\ \texttt{ff-ft} is superior to strong models such as \cite{dyer2016recurrent}, and it outperforms traditional parsers that used dozens of hand-crafted features to explicitly give the model structural information, such as \citet{zhu2013fast}'s parser.
Against other sequence tagging constituent parsers, we improve by more than two points the 91.2\% by \citet{VilaresMTL2019}, which was the current state of the art for this particular paradigm, and used task-specific decoders and multi-task learning, together with PoS tags and character embeddings. For dependencies, we observe similar tendencies. With respect to the state of the art, the winning system \cite{che2018towards} at the CoNLL-UD shared task 
reported a  \textsc{las} of 84.57\% with an ensemble approach that incorporated \elmo\ vectors (and also additional features such as PoS tag embeddings). \bert\  \texttt{ff-ft} also performs close to widely used approaches such as \cite{kiperwasser2016simple}. When comparing against sequence labeling dependency parsers, we obtain a \textsc{las} of 78.8\% versus the 78.6\% reported by \cite{StrzyzViable2019}, which included task-specific decoders and linguistic features, such as PoS tags and character embeddings.

\begin{table}[hbtp]
\centering
\small{
\begin{tabular}{lc}
\hline
\bf \multirow{2}{*}{Models} & \bf \textsc{ptb} \\
&\bf \textsc{f1}\\
\hline
\texttt{ff}$_{\bert}$&78.2\\
\texttt{ff-ft}$_{\bert}$&93.5\\
\hline
\citet{vinyals2015grammar}&88.3\\
\citet{gomez2018constituent}$^\diamond$&90.7\\
\citet{zhu2013fast}&90.4\\
\citet{VilaresMTL2019}$^\diamond$&91.2\\
\citet{dyer2016recurrent} (generative)&92.1\\
\citet{kitaev2018constituency}&95.1\\
\citet{kitaev2018BERT}&\bf95.8\\
\citet{zhou-zhao-2019-head}&\bf 95.8\\
\hline

\end{tabular}}
\caption{Comparison against related work on the \textsc{ptb}. 
$\diamond$ are other sequence tagging parsers.}\label{table-SOTA-const}
\end{table}

\begin{table}[hbtp]
\tabcolsep=0.10cm
\centering
\small{
\begin{tabular}{lcc}
\hline
\bf \multirow{2}{*}{Models} & 
\multicolumn{2}{c}{\bf \textsc{en-ewt}} \\
&\bf \textsc{las}&\bf \textsc{uas}\\
\hline
\texttt{ff}$_{\bert}$&63.0&68.1\\
\texttt{ff-ft}$_{\bert}$&78.8&81.0\\
\hline
\citet{StrzyzViable2019}$^\diamond$&78.6&81.5\\
\citet{kiperwasser2016simple}&79.0&82.2\\
\citet{straka:2018:K18-2} (UDpipe 2.0)&82.5&85.0\\
\citet{che2018towards}&\bf 84.6&\bf 86.8\\
\hline

\end{tabular}
}
\caption{Comparison against related work on the \textsc{en-ewt ud} test set. $\diamond$ are other sequence tagging parsers.}\label{table-SOTA-dep}
\end{table}

\subsection{Analysis of syntactic abilities}

Here, we consider the \texttt{ff} models for two reasons: they (i) do not fine-tune word vectors, and (ii) have limited expression.

\begin{figure}[t]
\centering
\includegraphics[width=1.\columnwidth]{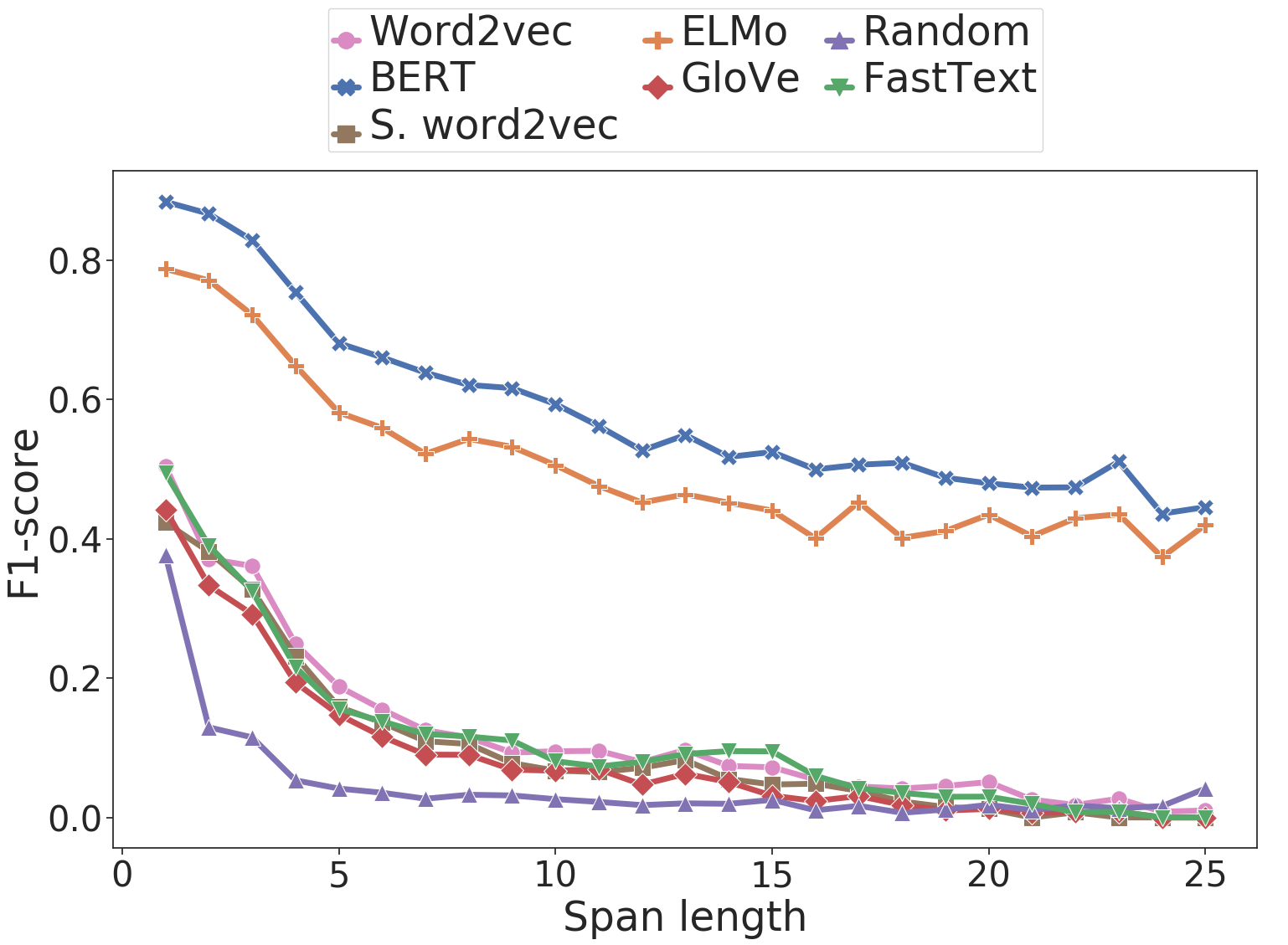}
\caption{\label{f-seq-lab-results} Span length F1-score on the \textsc{ptb} test set for the \texttt{ff} models}
\end{figure}

\begin{figure}[t]
\centering
\includegraphics[width=1.0\columnwidth]{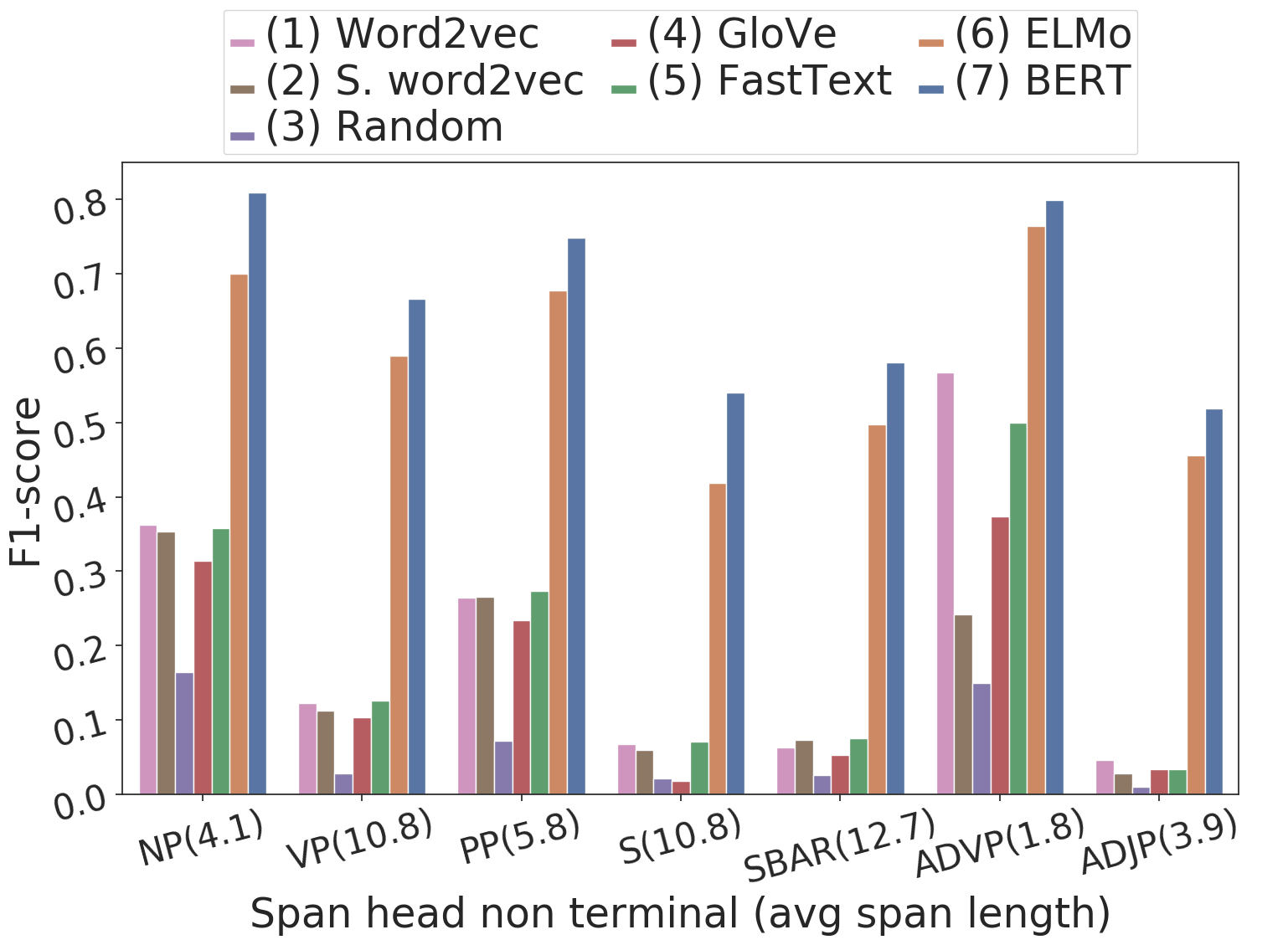}
\caption{Span label F1-score on the \textsc{ptb} test set for the \texttt{ff} models\label{f-seq-lab-non-terminal-results}. }
\end{figure}

Figure \ref{f-seq-lab-results} shows the F1-score on span identification for different lengths. All precomputed vectors surpass the random model. For contextualized vectors, although \bert\ does consistently better than \elmo, both  fairly succeed at identifying labeled spans. 
Figure \ref{f-seq-lab-non-terminal-results} shows the performance by span label.
Labels coming from shorter and frequent spans (e.g. \textsc{np}) are easier to handle by precomputed vectors. However, these struggle much more than contextualized ones when these come from larger or less frequent spans (e.g. \textsc{vp}).\footnote{
For unary chains, we only pick up the uppermost element. Also, the span containing the whole sentence is ignored.}

\begin{figure}[t]
\centering
\includegraphics[width=1.0\columnwidth]{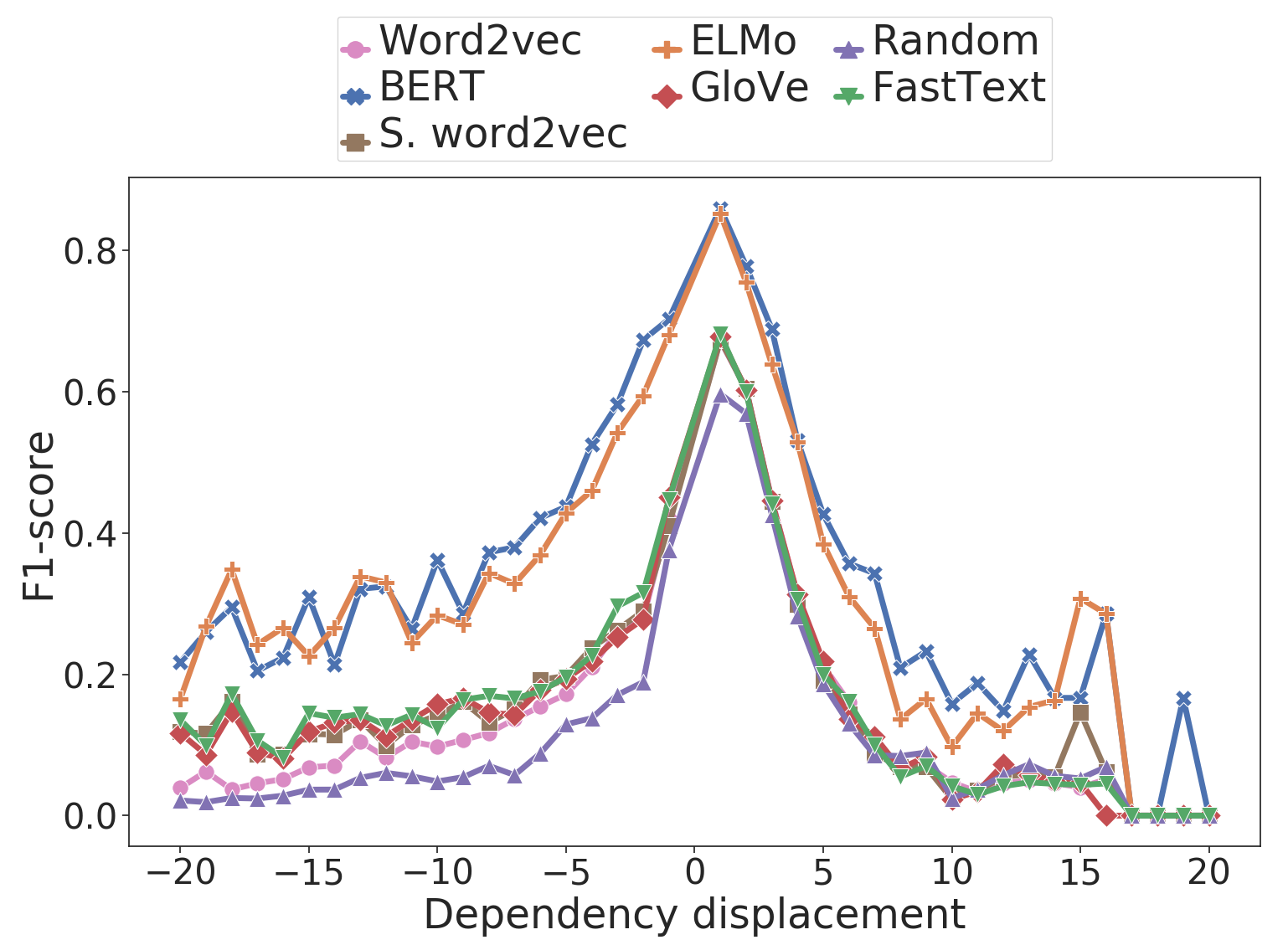}
\caption{\label{f-dependency-displacement-example}  Dependency displacement F1-score on the \textsc{en-ewt ud} test set for the \texttt{ff} models (with gold segmentation)}
\end{figure}

\begin{figure}[t]
\centering
\includegraphics[width=1.0\columnwidth]{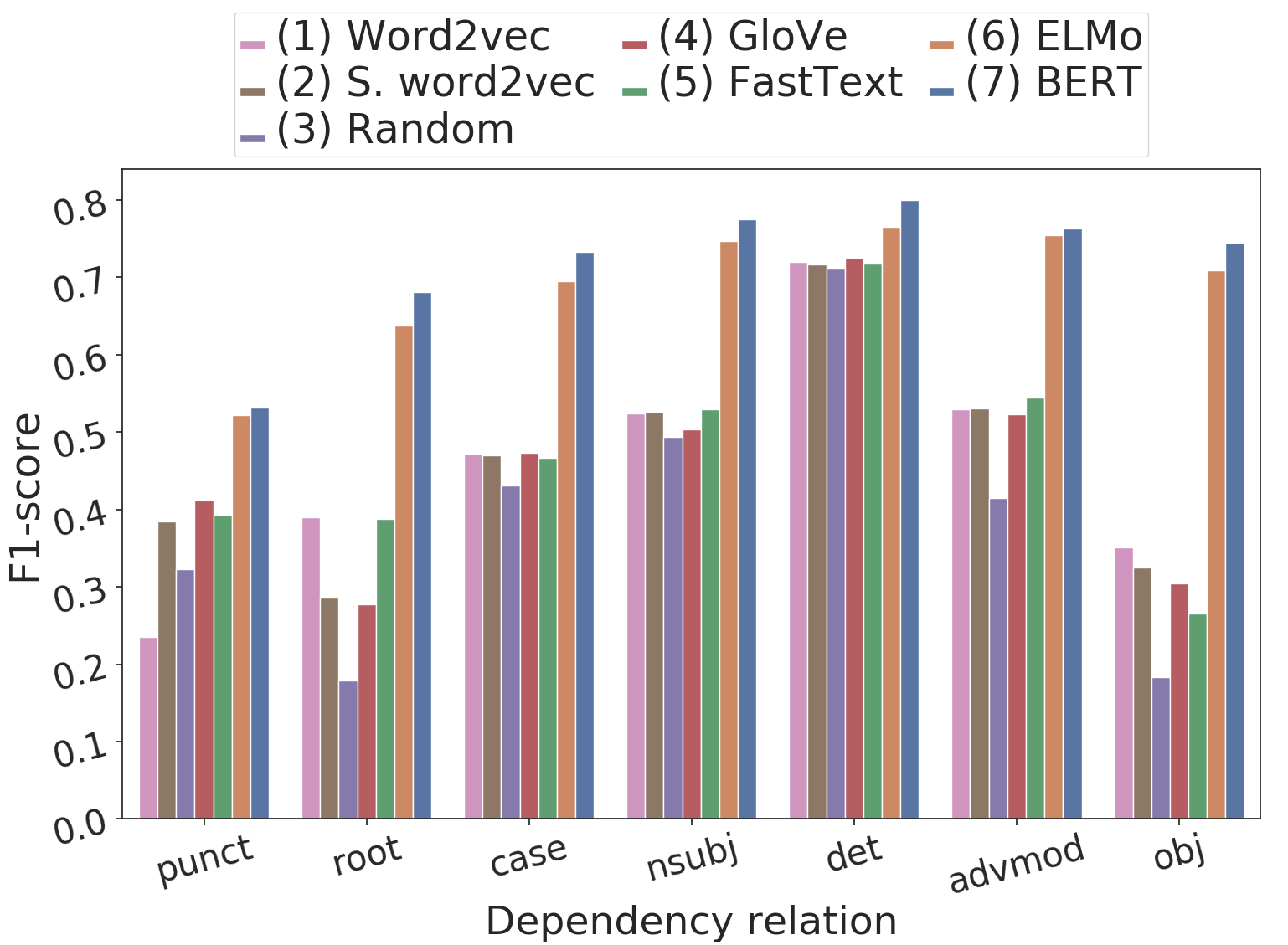}
\caption{F1-score for the most common relations 
on the \textsc{en-ewt ud} test set for \texttt{ff} models (with gold segmentation)\label{f-seq-lab-deprel-results}}
\end{figure}

Figure \ref{f-dependency-displacement-example} shows the analysis for dependencies, using dependency displacements. The results for precomputed embeddings are similar across positive and negative displacements. For contextualized vectors, the differences between both \elmo\ and \bert\ are small, and smaller than for constituents. They also perform closer to precomputed vectors, suggesting that these models could be less suited for dependency structures. The results also show similarities to the analysis on the constituent and dependency subtasks proposed by \citet{tenney2018you}, who expose that \elmo\ and \bert\ perform closer to each other when they are asked to find out dependency relations. 
Figure \ref{f-seq-lab-deprel-results} shows the performance on common dependency relations. Simple relations such as \texttt{det} are handled accurately by both precomputed and contextualized vectors,\footnote{Other simple relations such as \texttt{punct} or \texttt{case} also showed a high performance, but only when evaluating relation labels only (i.e. when failing the head index was not penalized, as opposed to Figure \ref{f-seq-lab-deprel-results}). These results are not included due to space limitations.} while harder ones such as \texttt{root}, \texttt{nsubj} or \texttt{obj} need deep contexualized ones.

In general, we feel Figures \ref{f-seq-lab-results}-\ref{f-seq-lab-deprel-results} distill that \elmo\ and \bert\ representations respond better to phrase-structure representations. Our intuition is that this might be due to language modelling objectives, where learning the concept of `phrase' seems more natural that the one of `dependency', although the answer to this is left as an open question.

\section{Conclusion}

We proposed a method to do constituent and dependency parsing relying solely on pretraining architectures -- that is, without defining any parsing algorithm or task-specific decoders.
Our goal was twofold: (i) to show to what extent it is possible to do parsing relying only in word vectors, and  (ii) to study if certain linguistic structures are learned in pretraining networks. To do so, we first cast parsing as sequence labeling, to then map; through a linear layer, words into a sequence of labels that represent a tree. During training, we considered to both freeze and fine-tune the pretraining networks. The results showed that (frozen) pretraining architectures such as \elmo\ and \bert\ get a sense of the syntactic structures, and that a (tuned) \bert\ model suffices to parse. 
Also, by freezing the weights we have provided different analyses regarding the syntax-sensitivity of word vectors.

Contemporaneously to this work, \citet{hewitt-liang-2019-designing} proposed to complement probing frameworks that test linguistic abilities of pretrained encoders with \emph{control tasks}, i.e. tasks that can be only learned by the probing framework (e.g. classifying words into random categories). If the pretrained network is encoding the target property, the probing framework should perform well on the target task and poorly on the control one. As future work, we plan to add this strategy to our analyses, and expand our experiments to languages other than English.

\section*{Acknowledgements}

We thank Mark Anderson and Daniel Hershcovich for their comments. DV, MS and CGR are funded by the ERC under the European Union's Horizon 2020 research and innovation programme (FASTPARSE, grant No 714150), by the
ANSWER-ASAP project (TIN2017-85160-C2-1-R) from MINECO, and by Xunta de Galicia (ED431B 2017/01). AS is funded by a Google Focused Research Award.

\bibliography{AAAI-VilaresD.2087}
\bibliographystyle{aaai}

\end{document}